\documentclass[11pt]{article}

\usepackage[utf8]{inputenc}
\usepackage[english]{babel}
\usepackage{enumerate}

\usepackage[dvipsnames]{xcolor}
\usepackage{xcolor}

\usepackage[margin=3.5cm]{geometry}

\usepackage{authblk}

\usepackage{graphicx}
\graphicspath{ {./img/} }
\usepackage{float}  
\usepackage{placeins}  

\usepackage{natbib}
\usepackage{breakcites}
\usepackage{amsmath}
\usepackage{amsthm}
\usepackage{amssymb}
\usepackage{amsfonts}

\usepackage{hyperref}
\hypersetup{hyperindex,breaklinks}

\usepackage{glossaries}
\newacronym{or}{OR}{operations research}
\newacronym{co}{CO}{combinatorial optimization}
\newacronym{ml}{ML}{machine learning}
\newacronym{dl}{DL}{deep learning}
\newacronym{rl}{RL}{reinforcement learning}

\newacronym{iid}{iid}{independent and identically distributed}
\newacronym{sgd}{SGD}{stochastic gradient descent}
\newacronym{mdp}{MDP}{Markov decision process}

\newacronym{nn}{NN}{neural network}
\newacronym{mlp}{MLP}{multilayer perceptron}
\newacronym{cnn}{CNN}{convolutional neural network}
\newacronym{rnn}{RNN}{recurrent neural network}
\newacronym{gnn}{GNN}{graph neural network}
\newacronym{gan}{GAN}{generative adversarial network}

\newacronym{tsp}{TSP}{traveling salesman problem}
\newacronym{lp}{LP}{linear programming}
\newacronym{milp}{MILP}{mixed-integer linear programming}
\newacronym{qp}{QP}{quadratic programming}
\newacronym{miqp}{MIQP}{mixed-integer quadratic programming}
\newacronym{sdp}{SDP}{semidefinite programming}
\newacronym{bnb}{B\&B}{branch-and-bound}
\newacronym{sat}{SAT}{boolean satisfiability problem}

\newacronym{gpu}{GPU}{graphical processing unit}
\newacronym{api}{API}{application programming interface}
\newacronym{gil}{GIL}{global interpreter lock}

\usepackage{listings}
\newcommand{\inline}[1]{\lstinline{#1}}
\definecolor{codegreen}{rgb}{0,0.6,0}
\lstdefinestyle{mystyle}{
    keywordstyle=\bfseries\color{codegreen},
    stringstyle=\color{green},
    basicstyle=\ttfamily\footnotesize,
    breakatwhitespace=false,
    breaklines=false,
    captionpos=b,
    keepspaces=true,
    frame=single,
    numbersep=5pt,
    showspaces=false,
    showstringspaces=false,
    showtabs=false,
    tabsize=2
}
\usepackage{csquotes}

\title{Ecole: A Library for Learning Inside MILP Solvers}

\author[1,2]{Antoine~Prouvost}
\author[1]{Justin~Dumouchelle}
\author[1,2]{Maxime~Gasse}
\author[1]{Didier~Chételat}
\author[1,2]{Andrea~Lodi}
\affil[ ]{\texttt{\{firstname\}.\{lastname\}@polymtl.ca}}
\affil[1]{Canada Excellence Research Chair in Data Science for Decision Making, École Polytechnique de Montréal}
\affil[2]{Mila, Quebec Artificial Intelligence Institute}
\date{\vspace{-5ex}}

\begin{document}
\maketitle

\begin{abstract}
    In this paper we describe Ecole (Extensible Combinatorial Optimization Learning Environments), a library to facilitate integration of machine learning in combinatorial optimization solvers.
    It exposes sequential decision making that must be performed in the process of solving as Markov decision processes.
    This means that, rather than trying to predict solutions to combinatorial optimization problems directly,
    Ecole allows machine learning to work in cooperation with a state-of-the-art a mixed-integer 
    linear programming solver that acts as a controllable algorithm.
    Ecole provides a collection of computationally efficient, ready to use learning
    environments, which are also easy to extend to define novel training tasks.
    Documentation and code can be found at \url{https://www.ecole.ai}.
\end{abstract}

\section{Introduction}
    Combinatorial optimization algorithms play a crucial role in our societies, for tackling a wide range of
    decision problems arising in, but not limited to, transportation, supply
	chain, energy, finance and scheduling \citep{paschos2013applications}.
    These optimization problems, framed as mathematical programs, are inherently hard to solve, forcing practitioners to
    constantly develop and improve existing algorithms.
    As a result, general-purpose mathematical solvers typically rely on a large number of handcrafted heuristics that are critical to efficient problem solving, but whose
    interplay is usually not well understood and is exponentially hard to analyse.
    These heuristics can be regarded as having been \emph{learned} by
    human experts through trial and error, on public (or private) data-sets of problems such as MIPLIB \citep{miplib2017}.

    At the same time, traditional solvers typically disregard the fact that some
    applications require to solve similar problems repeatedly, and tackle each new problem independently, without leveraging any knowledge from the past.
    In this context, applying \gls{ml} to \gls{co} appears as a natural idea, and has actually been a topic of interest for quite some time
    \citep{SmithNeuralNetworksCombinatorial1999}.
    With the recent success of \gls{ml}, especially the \gls{dl} sub-field, there
    is renewed appeal to replace some of the heuristic rules inside traditional solvers by statistical models learned from data.
    The result would be a solver whose performance could be automatically tailored to a given
    distribution of mathematical optimization problems, which could be either application-specific or general-purpose ones.
    The reader is referred to \cite{bengio2020machine} for a detailed survey on the topic.
    
    In this article, we present Ecole, an open-source library aimed at facilitating the development of \gls{ml} approaches within general-purpose \gls{milp} solvers based on the \gls{bnb} algorithm.
    The remainder of this article is organized as follows.
    In Section~\ref{sec:motivation}, we detail the challenges faced by practitioners for applying
    \gls{ml} inside \acrfull{co} solvers.
    In Section~\ref{sec:related_works}, we present existing software that also aim at facilitating the development of \gls{ml} solutions for \gls{co}.
    In Section~\ref{sec:background}, we provide background on \acrlong{milp} and the \acrlong{bnb}
    algorithm, as well as the concepts of \acrlong{mdp} and \acrlong{rl}.
    In Section~\ref{sec:solver_control_problem}, we present our formulation of control
    problems arising in mathematical solvers as Markov decision processes,
    and in Section~\ref{sec:design}, we showcase the Ecole interface and how it relates to this formulation.
    In Section~\ref{sec:experiments}, we compare the computing performance of Ecole for extracting solver features,
    compared to existing implementations from the literature.
    Finally, we conclude with a discussion on future plans for Ecole in Section~\ref{sec:future}.

\section{Motivation}
    \label{sec:motivation}
    Building the appropriate software to apply \gls{ml} inside of a \gls{bnb} solver is not an easy task, and requires a deep knowledge of the solver.
    It may take months of software engineering before researchers can focus on the actual \gls{ml} algorithm, and the engineering endeavors can be dissuasive.
    For example, it suffices to look at research articles with public software implementation
    \citep{conf/nips/HeDE14,gasse2019exact,gupta2020hybrid,zarpellon2020parameterizing} to get an idea of the complexity of the required code base.
    Not to mention the fact that such implementations can themselves contain bugs, and will quickly become outdated.

    Solvers such as
    SCIP \citep{scip}, CPLEX \citep{cplex}, and Gurobi \citep{gurobi}, expose their \gls{api} in the \inline{C} programming language, while the state-of-the-art
    tools for \gls{ml} such as Scikit-Learn \citep{scikit-learn},
    PyTorch \citep{pytorch}, and TensorFlow \citep{tensorflow} exist primarily
    in Python. Advanced software engineering skills are necessary to interface both ecosystems, and the room for errors is large, especially if additional time is not invested to
    write tests for the code.
    Once these hardships are overcome, the resulting implementation may still be slow and
    lack advanced features such as parallelization (in particular due to the Python \gls{gil} 
    that prevents multi-threaded code executions).
    Furthermore, research software written for particular projects is often difficult to reuse
    without copy-editing code, as they lack extensible concept abstractions, proper software packaging, and code maintenance.

    Ecole is a free and open-source library built around the SCIP solver to address the aforementioned
    issues.
    Several decision problems of interest that arise inside the solver are exposed through an extensible interface
    akin to OpenAI Gym library \citep{openaigym}, a library familiar to \gls{ml} practitionners.
    Going further, Ecole aims at improving the reproducibility of scientific research in the area with unified problem benchmarks and metrics, and provides strong default options for new researchers to start with, without the need for an expert knowledge of the inner workings of a mathematical solvers.

\section{Related Work}
    \label{sec:related_works}
    Other libraries have been introduced recently to facilitate the application of \gls{ml} to operations research.
    MipLearn \citep{miplearn} is basically aimed at the same goals as Ecole, with a strong focus on customization and extensibility. It supports two competitive commercial solvers, namely CPLEX and
    Gurobi, but as a result is limited in the type of interactions it offers, and only allows for using \gls{ml} for solver configuration. In contrast, Ecole only supports the open-source solver SCIP, but allows for repeated decision making, such the selection of branching variables during \gls{bnb}, which is a cornerstone of the algorithm.
    ORGym \citep{orgym} and OpenGraphGym \citep{opengraphgym} also offer Gym-like learning environments, for general \gls{or} problems and for graph-based problems, respectively. Both are aimed at
    using \gls{ml} to produce feasible solutions directly, without the need for an \gls{milp} solver.
    As such they do not allow for the exact solving of \gls{co} problems.
    Ecole, on the other hand, benefits from the inherent mathematical guarantees of a mathematical solver, which include the possibility of exact solving.
    As such Ecole does not necessarily offer a replacement to the existing software in the \gls{ml} for \gls{or} ecosystem, but rather a (nice) complement that fills some existing gaps.
    For instance, practitioners can use one of the problem benchmarks from MipLearn, ORGym or OpenGraphGym, to generate a collection of instances in a standard format, and then use Ecole for learning to branch via \gls{ml}.

\section{Background}
    We now introduce formally some key concepts that are relevant for describing Ecole, related to both combinatorial optimization and \acrlong{rl}.

    \label{sec:background}
    \subsection{Combinatorial Optimization}
        \label{sec:background:co}
        Mathematical optimization can be used to model a variety of problems.
        Variables model the decisions to be made, constraints represent physical or structural
        restrictions, while the objective function defines a measure of cost to be minimized.
        When the objective and constraints are linear, and some variables are restricted to be integer,
        the problem is a \acrfull{milp} problem.
        \gls{milp} is an important class of decision problems but \gls{milp} problems are, in general,
        $\mathcal{NP}$-hard to solve.
        The \gls{bnb} algorithm \citep{land1960automatic} is an implicit enumeration scheme that is
        generally at the core of the mathematical solvers designed to tackle these problems.
        The algorithm starts by computing the (continuous) \gls{lp} relaxation, typically using the
        Simplex algorithm \citep{dantzig1990origins}.
        If the solution respects the integrality constraints, then the solution is optimal.
        Otherwise, the feasible space is split by branching on (\textit{i.e.}, partitioning the domain of) an integer
        variable in a way that excludes the solution to the current \gls{lp} relaxation.
        The algorithm is then recursively applied to the sub-domain.
        If a sub-domain \gls{lp} relaxation is infeasible, or if its objective value does not
        improve on the best feasible solution, then the algorithm can stop exploring it.

        General-purpose solvers, such as SCIP \citep{scip}, CPLEX \citep{cplex},
        and Gurobi \citep{gurobi}, have emerged to provide an enhanced version of the \gls{bnb} algorithm.
        They include additional techniques such as presolving \citep{achterberg2020presolve},
        cutting planes \citep{gomory2010outline,balas1993lift},
        and primal heuristics \citep{fischetti2011heuristics}, which together with \gls{bnb} have contributed
        to drastically reduce the solving time of \gls{milp} in the last decades \citep{bixby2007progress}.

    \subsection{Reinforcement Learning}
        \label{sec:background:rl}
\begin{figure}[H]
            \centering
            \includegraphics[width=.6\columnwidth]{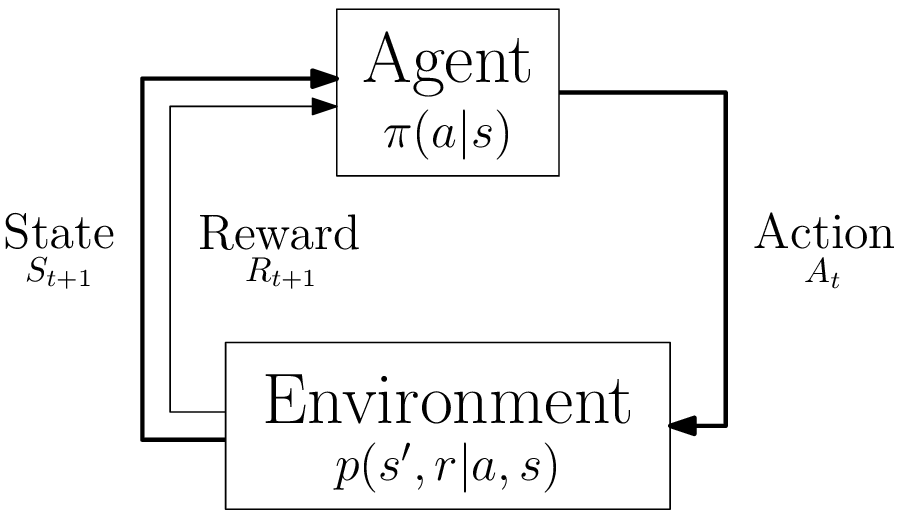}
            \caption{
                The \gls{mdp} associated with \acrlong{rl},
        		modified from \cite{sutton2018reinforcement}.
            }
            \label{fig:mdp}
        \end{figure}

        In \gls{rl}, an agent interacts with an environment and receive rewards as a (indirect)
        consequence of its actions.
        The frameworks is defined by \gls{mdp} problems as follows.
        At every time step $t$, the agent is in a given state $S_t$ and decides on an
        action $A_t$.
        The agent decisions are modeled by a probabilistic policy function:
        for a given state $s$ that the agent is in, the agent takes an action $a$ with probability
        $\pi(a|s)$.
        As a result, and depending on the unknown dynamics of the environment,
        the agent transitions into a new state $S_{t+1}$ and receives a reward $R_{t+1}$.
        If the agent is in a state $s$, and takes an action $a$, then the probability to
        transition into a new state $s'$ and receiving a reward $r$ is denoted by
        $\mathbb{P}(s',r|a,s)$.
        This is illustrated in Figure~\ref{fig:mdp}.
        The process ends when reaching a state defined to be terminal, where the set
        of possible actions is empty.
        The objective for the agent is to maximize the expected sum of future rewards.

        A sequence of actions and transitions $\tau$ is called an \emph{episode},
        or \emph{trajectory}.
        The probability of a given trajectory is given by 

        \begin{equation}
        \label{eq:prob_traj}
            \mathbb{P}(\tau) =
            \mathbb{P}(S_0)
            \prod_t \pi(A_t|S_t)\mathbb{P}(S_{t+1},R_{t+1}|A_t,S_t).
        \end{equation}

        Common \gls{rl} algorithms fall in two categories.
        Policy methods try to directly learn an approximation of the agent policy function $\pi$.
        Value based methods estimate the state-action value function $Q(s, a)$ defined as the
        expected sum of future rewards given that the agent is in a state $s$ and takes an action $a$.
        The associated policy is then computed through approximate dynamic programming
        \citep{bertsekas2008approximate}.
        Finally, modern approaches such as actor-critic methods both train a policy (the actor) and an
        estimate of the value function of the policy (the critic) in a single procedure.

    \subsection{Imitation Learning}
        An alternative to find good policies in a \gls{mdp} is to learn to imitate another
        expert policy, an approach usually referred to as imitation learning.
        In this scenario, something usually prevents using the expert directly: for example, the expert
        might be a human, or might be an expensive algorithm that takes excellent decisions at high
        computational cost.

        The simplest possible approach for imitation learning is behavioral cloning, where state-action
        pairs are collected by running the expert for a while, and a \gls{ml} model is trained by imitation
        learning to predict actions that would be taken by the expert in those states.
        A downside of this approach is the state distribution mismatch problem: as the student will usually
        make mistakes, it will deviate from the kinds of states likely to be encountered by the expert
        and will soon end up in new states much unlike those seen during training.
        So called active imitation learning methods, such as DAGGER \citep{ross2011reduction}, try to
        correct this mismatch to improve final policy performance.

\section{Solver Control Problem}
    \label{sec:solver_control_problem}
    Although \gls{milp} solvers are usually deterministic processes, in Ecole we adopt a general
    \gls{mdp} formulation, introduced in Section~\ref{sec:background:rl}, which allows for
    non-deterministic processes.

    To better map the Ecole interface presented in the next section, we further refine
    probability \eqref{eq:prob_traj}.
    First, we split the initial state distribution $\mathbb{P}(S_0)$ using Bayes' rule to introduce
    the probability distribution of the problem instance $I$ being solved, which can be formulated
    as $\mathbb{P}(S_0) = \mathbb{P}(I)\mathbb{P}(S_0|I)$.
    Second, we borrow from the partially-observable \gls{mdp} framework \citep{spaan2012partially}
    and introduce an observation function $\mathcal{O}$ of the state that is used by the agent to
    make decisions, using all of the past trajectory, that is
    $\pi(A_t|\mathcal{O}(S_0),...,\mathcal{O}(S_t))$.
    Similarly, we reformulate rewards as a function $\mathcal{R}$ of the state and leave it out of
    the transition probability $\mathbb{P}(S_{t+1}|A_t,S_t)$.
    This gives us the following final probability distribution of a trajectory:

    \begin{equation}
    \label{eq:prob_traj_ecole}
        \mathbb{P}(\tau) =
        \mathbb{P}(I)\mathbb{P}(S_0|I)
        \prod_t \pi(A_t|H_t) \;
        \mathbb{P}(S_{t+1}|A_t,S_t),
    \end{equation}
    where the history $H_t$ is given by
    \begin{equation}
        H_t = \{
            \mathcal{O}(S_0),\mathcal{R}(S_0), A_0,...,
            \mathcal{O}(S_{t-1}),\mathcal{R}(S_{t-1}), A_{t-1}, \mathcal{O}(S_t)
        \}.
    \end{equation}

    Ecole provides an \gls{api} to define environments inside the solver.
    It lets users sample from those environments according to~\eqref{eq:prob_traj_ecole},
    as well as extract rewards and observations using the functions $\mathcal{R}$ and $\mathcal{O}$.
    Two environments are currently implemented in Ecole.

    \paragraph{Configuring}
        The first environment expose the task of algorithm configuration
        \citep{hoos2011automated}.
        The goal is to find well performing SCIP parameters, then let the solver run its course
        without intervention.
        This task is akin to what is learned in \cite{hutter2010automated}.
        In this scenario, finding optimal parameters can be framed as a contextual bandit problem\footnote{
            A good reference on the topic is \citet[Part V]{lattimore2020bandit}.
        }.
        Contextual bandit problems are special cases of \gls{rl} where the underlying \gls{mdp} has unit episode
        length, so it fits naturally in the framework adopted by Ecole as a special case.
        In this environment, the one and only action to be taken is a mapping of SCIP parameters and associated values.

    \paragraph{Branching}
        The second environment implemented allows users to select the variable to branch on each
        \gls{bnb} node, as used in \cite{gasse2019exact}.
        The state $S_t$ is defined as the state of the solver on the $t$\textsuperscript{th} node,
        and is equivalent to the branching rule callback available in SCIP.
        In \cite{gasse2019exact}, the observation function $\mathcal{O}$ is extracting a
        bipartite graph representation of the solver, and $\pi$ is a \gls{gnn}.

    Plans for future environments are suggested in Section~\ref{sec:future}.

\section{Design Decisions}
\label{sec:design}
    The \gls{api} of Ecole is designed for ease of use and extensibility.
    In this section, we detail some of the key features of Ecole.

    \subsection{Environment Interface}
        \label{sec:environment_api}
        \paragraph{Environments}
        The interface for using an environment is inspired by the OpenAI Gym \citep{openaigym} library.
        The main abstraction is the \inline{Environment} class, which is used to encapsulate any
        control problem, as formulated in Section~\ref{sec:solver_control_problem}.
        The Listing~\ref{lst:ecole_api} provides an example of using Ecole with the
        \inline{Branching} environment for \gls{bnb} variable selection.
        The inner \inline{while} loop spans over a full episode, while the outer
        \inline{for} loop repeats it multiple times for different problem instances.
        An episode, starts with a call to \inline{reset}.
        The method takes as parameter an \gls{milp} problem instance from which an initial state will be sampled,
        and returns an observation of that state, and a Boolean flag indicating whether that state
        is terminal.
        Transitions are performed by calling the \inline{step} method, with the action provided
        by the user (the \inline{policy} function).
        It returns the observation of the new state, the reward, the flag indicating
        whether the new state is terminal, and a dictionary of additional non-essential
        information about the transition.

        \bigskip
        \centerline{\begin{minipage}{1.05\linewidth}
\begin{lstlisting}[
            label=lst:ecole_api,
            caption=Default usage of environments in Python.,
            language=Python
        ]
import ecole

env = ecole.environment.Branching(
    reward_function=ecole.reward.LpIterations() ** 2,
    observation_function=ecole.observation.NodeBipartite(),
)
instances = ecole.instance.IndependentSetGenerator(n_nodes=100)

for _ in range(10):
    obs, action_set, reward_offset, done, info = env.reset(next(instances))
    while not done:
        action = policy(obs, action_set)
        obs, action_set, reward, done, info = env.step(action)
\end{lstlisting}
        \end{minipage}}

        \paragraph{Reward and Observation Functions}
        Furthermore, Listing~\ref{lst:ecole_api} shows how the constructor of \inline{Environment}
        can be used to specify the rewards and observations to compute.
        Some observation functions from the literature are provided Ecole, such as the ones used in
        \cite{gasse2019exact} and \cite{khalil2016learning}.
        The listing also demonstrate how new reward functions can be dynamically created by applying
        mathematical operations (\inline{+}, \inline{-}, \inline{*}, \inline{/}, \inline{**}, \inline{.exp()}...)
        on them.

        \paragraph{Instance Generators}
        Learning inside a solver may require large amount of training instances.
        Although industry applications can provide extensive datasets of instances of interest, it is also
        valuable to have on hand generators with good defaults to quickly experiment ideas.
        Out of convenience, we provide four families of generators for users to learn from.
        These are the same families of instances that were used to benchmark the imitation learning method of
        \citet{gasse2019exact}.
        \begin{itemize}
            \item Set covering \gls{milp} problems generated following \citep{balas1980set};
            \item Combinatorial auction \gls{milp} problems generated following
                \cite[][Section 4.3]{leyton2000towards};
            \item Capacitated facility location \gls{milp} problems generated following
                \citep{cornuejols1991comparison};
            \item Independent set \gls{milp} problems generated following the procedure of
                \cite[][Section 4.6.4]{bergman2016decision} with both Erdos-Renyi \citep{erdos1959random}
                and Barabasi-Albert \citep{barabasi1999emergence} graphs.
        \end{itemize}

        Table~\ref{table:ecole_api} summarizes the key abstractions in Ecole and how they map to the
        mathematical formulation presented in Section~\ref{sec:solver_control_problem}.
        Some elements are further explained in the next section.

        \begin{table}[H]
        \centering
        \begin{tabular}{ |c|c|c| }
            \hline
            Observation function & $\mathcal{O}$ & \inline{NodeBipartite()} \\
            Reward function & $\mathcal{R}$ & \inline{LpIterations() ** 2} \\
            Instance dist. & $\mathbb{P}(I)$ & \inline{IndependentSetGenerator(n_nodes=100)} \\
            State & $S_t$ & \inline{Model} \\
            Cond. initial state dist. & $\mathbb{P}(S_0|I)$ & \inline{BranchingDynamics.reset_dynamics} \\
            Policy & $\pi(A_t|H_t)$ & \inline{policy} \\
            Transition dist. & $\mathbb{P}(S_{t+1}|A_t,S_t)$ & \inline{BranchingDynamics.step_dynamics}\\
            \hline
        \end{tabular}
        \caption{Comparing Ecole \gls{api} to its mathematical formulation.}
        \label{table:ecole_api}
        \end{table}

    \subsection{Extensibility}
    \label{sec:extensibility}
        OpenAI Gym is designed as a set of benchmarks for \gls{ml} practitioners.
        However, in Ecole the tasks also have industry applications, and therefore it was an essential
        design principle to allow users flexibility in designing the environments.
        For example, although good defaults bring value, if users decide to train an agent with a customized
        observations or reward function and this leads in the end to better solving times, this is a net
        gain and Ecole should allow for it.
        Thus, environments were made to be customizable, unlike in OpenAI Gym.

        In this section, we explain how users can customize the reward
        function $\mathcal{R}$, the observation function $\mathcal{O}$, the instance distribution
        $\mathbb{P}(I)$, and the transition dynamics $\mathbb{P}(S_0|I)$ and
        $\mathbb{P}(S_{t+1}|A_t,S_t)$.

        \paragraph{Reward and Observation Functions}
        Users can create observation or reward functions by creating a class with two methods, as shown in
        Listing~\ref{lst:obs_func}.
        They have access to the state of the \gls{mdp}, \textit{i.e.}, the underlying SCIP solver,
        through the \inline{ecole.scip.Model}, or equivalently a \inline{pyscipopt.Model} object
        (both being wrappers of a \inline{SCIP*} pointer in \inline{C}).
        There is no limitation to what an observation can be because they are an abstraction used exclusively
        by the user.

        \bigskip
        \centerline{\begin{minipage}{1.05\linewidth}
\begin{lstlisting}[
            label=lst:obs_func,
            caption=Python interface of an observation function.,
            language=Python
        ]
class Observation:
    ...

class ObservationFunction:
    def before_reset(model: ecole.scip.Model) -> Observation:
        ...

    def extract(model: ecole.scip.Model) -> Observation:
        ...
\end{lstlisting}
        \end{minipage}}

        \paragraph{Transitions and Initial State}
        The initial state and transition probability distribution can be customized by creating a \inline{EnvironmentDynamics} object.
        Its \gls{api} is similar to that of the \inline{Environment}, with the exception that
        the two methods \inline{reset_dynamics} and \inline{step_dynamics} solely manipulate the
        solver, without computing either observations or rewards.
        Environments are actually wrappers around dynamics that also call the reward and observation
        functions.

        \paragraph{Instance Generators}
        Instance generators are regular Python generators that output an \inline{ecole.scip.Model}.
        The users are free to define any new generators that can create problem instances,
        or read them from file.

        For all the abstractions above, existing components of Ecole can easily be reused through
        composition or inheritance to speed up development.

    \subsection{Comparison with OpenAI Gym}
        One objective of developing Ecole is to provide an interface close to the popular
        OpenAI Gym library.
        Nonetheless, some differences exist.

        \paragraph{Condition Initial State distribution}
        The \inline{reset} method in Gym does not accept any parameters.
        In Ecole, it makes little sense to solve the same \gls{milp} instance over and over,
        hence conditioning the initial state probability distribution on the instance is mandatory.

        \paragraph{Initial Terminal States}
        In the Gym interface, initial states cannot be terminal.
        As a result the \inline{reset} method does not return the boolean termination flag.
        However, terminal initial states do arise in the environments defined in Ecole.
        For example, in \gls{bnb} variable selection, the instance could be solved through preprocessing 
        and never require any branching.

        \paragraph{Reward Offsets}
        In Ecole, rewards are also returned on \inline{reset}.
        This is because rewards usually come from differences of metrics, but users cannot compute the
        total metric without knowing how the metric evolved until the first decision point.
        The reset reward is this missing information.
        For example, the solving time reward reports how much time is spent between each decision and the next state.
        Summing these rewards give the total time spent solving since the agent was first asked for a decision,
        but it does not take into account time spent before that point.
        In \gls{bnb} variable selection, this time would include preprocessing, and root node cutting plane calculation
        and \gls{lp} solving.
        This pre-decision time is what is returned by reset, allowing it to be summed to the rest of the rewards to
        find the total solving time.

        \paragraph{Action Spaces}
        In OpenAI Gym, \inline{action_space} is a property of the environment class, that is, it does not change
        between episodes.
        In Ecole, not only the set of actions can change from one instance to the next, it can also change between transitions.
        For instance, in \gls{bnb} variable selection, the set of valid branching candidates changes on every node.

    \subsection{Ecosystem}
    \label{sec:ecosystem}
        The largest part of Ecole is written in \inline{C++}, and exposes bindings to Python
        through PyBind11 \citep{pybind11}.
        Xtensor \citep{xtensor} is used for high level multi-dimensional arrays, and NumPy
        \citep{numpy} is used for binding them to Python.

        Writing Ecole in \inline{C++} provides solid ground for a computationally
        efficient library that can be made available in multiple programming languages
        \citep{xtensor-vision}.

\section{Performance Experiments}
    \label{sec:experiments}
    Speed is a core principle of Ecole.
    Besides having most of its codebase in \inline{C++}, selecting carefully the third-party libraries on which
    it relies (see Section~\ref{sec:ecosystem}) and tuning compilation options, other optimizations were
    leveraged to further speedup Ecole.
    When using Ecole from Python, calls into the \inline{C++} code are made without copying
    parameters and return values unless necessary.
    Furthermore, any function doing significant work, such as data extraction or calls to the SCIP solver,
    are performed without holding the \gls{gil}, allowing Python users to use Python threading capabilities
    and avoid the overhead that comes with multiprocessing parallelization.

    We propose two experiments to illustrate the gains provided by these optimizations and benchmark Ecole
    performance.\footnote{
        The code of the experiments is available at \url{https://github.com/ds4dm/ecole-paper}.
    }
    The benchmarks were run on a server with Linux OpenSUSE Tumbleweed 20210114, with 32GB of RAM, and
    and Intel i7-6700K (8 cores, 4.0 GHz) CPU.
    To analyze the results, we used SciPy \citep{scipy}, Jupyter \citep{jupyter}, Pandas \citep{pandas},
    Matplotlib \citep{matplotlib}, and Seaborn \citep{seaborn}.

    \subsection{Ecole Overhead Experiment}
        The first concern was to understand whether using Ecole without extracting any data creates
        any overhead compared to using SCIP directly.
        In particular, SCIP offers the possibility to select a branching variable through the use of a
        callback function, while the Ecole interface presented in Section~\ref{sec:environment_api} wraps the
        callback function to select branching variables iteratively (effectively transforming the callback
        in a stackful coroutine, \textit{i.e.}, a function that can be suspended and resumed).

        Using the four instance generators from \citep{gasse2019exact} (implemented in Ecole and presented
        in Section~\ref{sec:extensibility}), we compare branching on the first available branching candidate
        using Ecole and vanilla SCIP.
        To speedup the benchmark, we disable presolving, cutting planes, and limit the number of nodes to 100.
        Over a total of $4500$ instances, a one sample t-test shows that the ratio of the wall times is not
        significantly different from $1.0$ (with $\text{p-value} < 10^{-50}$).
        Thus, we can conclude that for a typical usage, Ecole produces no overhead.
        This is explained by the fact that any possible overhead time is dwarfed in comparison to the time
        spent solving the problem.

    \subsection{Observation Functions Experiment}
        Our second experiment aims at measuring the sole execution time of two observation functions implemented
        in Ecole: \inline{NodeBipartite} from \cite{gasse2019exact} and \inline{Khalil2016} from
        \cite{khalil2016learning} and implemented in \cite{gasse2019exact}.
        Using the four instance generators from \cite{gasse2019exact}, with no presolving,
        no cutting planes, and limiting to 100 nodes, we measure their sole execution time.
        A total of 439 instances were generated.

        The results are given in Table~\ref{table:khalil2016} for \inline{Khalil2016}.
        For \inline{NodeBipartite}, it is possible to cache some information computed at the root node
        for future use but this is incompatible with cutting planes.
        The results without and with a cache are given in
        Tables~\ref{table:nodebipartite-nocache}~and~\ref{table:nodebipartite-cache}, respectively. 

        \begin{table}[H]
        \centering
        \begin{tabular}{|l|l|l|l|}
            \hline
            Generator & Wall time Gasse (s) & Wall time Ecole (s) & Ratio \\
            \hline
            CFLP 100-100 & $(1.067\pm0.072)*10^2$ & $1.050\pm0.117$ & $102.2\pm 5.9$ \\
            CFLP 200-100 & $(4.381\pm0.641)*10^2$ & $2.580\pm0.420$ & $170.2\pm 8.8$ \\
            CFLP 400-100 & $(2.468\pm0.354)*10^3$ & $6.151\pm0.768$ & $400.3\pm13.7$ \\
            \hline
            CAuction 100-500  & $(5.682\pm0.856)*10^{-1}$ & $(3.752\pm0.605)*10^{-2}$ & $15.19\pm0.77$ \\
            CAuction 200-1000 & $2.654\pm0.437$         & $(1.067\pm0.190)*10^{-1}$ & $24.93\pm0.74$ \\
            CAuction 300-1500 & $7.207\pm0.718$         & $(1.967\pm0.209)*10^{-1}$ & $36.68\pm0.98$ \\
            \hline
            IndependentSet 500  & $1.531\pm0.192$ & $(3.210\pm0.285)*10^{-1}$ & $4.755\pm0.212$ \\
            IndependentSet 1000 & $8.209\pm1.463$ & $1.424\pm0.146$ & $5.726\pm0.512$ \\
            IndependentSet 1500 & $(2.226\pm0.407)*10$ & $3.060\pm0.284$ & $7.228\pm0.724$ \\
            \hline
            SetCover 500-1000  & $(9.962\pm6.410)*10^{-2}$ & $(9.261\pm6.197)*10^{-3}$ & $10.26\pm 2.99$ \\
            SetCover 1000-1000 & $(4.351\pm2.812)*10^{-1}$ & $(5.136\pm3.403)*10^{-2}$ & $8.613\pm0.540$ \\
            SetCover 2000-1000 & $1.210\pm0.627$         & $(1.980\pm1.135)*10^{-1}$ & $6.274\pm0.555$ \\
            \hline
        \end{tabular}
        \caption{Comparison of execution times for \inline{Khalil2016}}.
        \label{table:khalil2016}
        \end{table}



        \begin{table}[H]
        \centering
        \begin{tabular}{|l|l|l|l|}
            \hline
            Generator & Wall time Gasse (s) & Wall time Ecole (s) & Ratio \\
            \hline
            CFLP 100-100 & $(4.232\pm0.070)*10^{-1}$ & $(1.209\pm0.038)*10^{-1}$ & $3.505\pm0.113$ \\
            CFLP 200-100 & $(8.017\pm0.384)*10^{-1}$ & $(2.815\pm0.125)*10^{-1}$ & $2.848\pm0.058$ \\
            CFLP 400-100 & $1.557\pm0.016$         & $(6.179\pm0.101)*10^{-1}$ & $2.520\pm0.037$ \\
            \hline
            CAuction 100-500  & $(7.376\pm0.549)*10^{-2}$ & $(4.074\pm0.348)*10^{-3}$ & $18.14\pm0.89$ \\
            CAuction 200-1000 & $(9.797\pm0.473)*10^{-2}$ & $(8.753\pm0.566)*10^{-3}$ & $11.21\pm0.36$ \\
            CAuction 300-1500 & $(1.160\pm0.016)*10^{-1}$ & $(1.356\pm0.047)*10^{-2}$ & $8.564\pm0.226$ \\
            \hline
            IndependentSet 500  & $(6.806\pm0.043)*10^{-1}$ & $(1.679\pm0.017)*10^{-1}$ & $4.053\pm0.028$ \\
            IndependentSet 1000 & $2.564\pm0.012$         & $(8.849\pm0.090)*10^{-1}$ & $2.898\pm0.026$ \\
            IndependentSet 1500 & $6.697\pm0.421$         & $2.046\pm0.015$         & $3.273\pm0.201$ \\
            \hline
            SetCover 500-1000  & $(4.347\pm2.556)*10^{-2}$ & $(3.986\pm2.413)*10^{-3}$ & $10.25\pm2.94 $ \\
            SetCover 1000-1000 & $(1.015\pm0.283)*10^{-1}$ & $(1.238\pm0.412)*10^{-2}$ & $8.368\pm0.749$ \\
            SetCover 2000-1000 & $(1.768\pm0.304)*10^{-1}$ & $(2.896\pm0.628)*10^{-2}$ & $6.179\pm0.469$ \\
            \hline
        \end{tabular}
        \caption{Comparison of execution times for \inline{NodeBipartite} without cache.}
        \label{table:nodebipartite-nocache}
        \end{table}
        
        \begin{table}[H]
        \centering
        \begin{tabular}{|l|l|l|l|}
            \hline
            Generator & Wall time Gasse (s) & Wall time Ecole (s) & Ratio \\
            \hline
            CFLP 100-100 & $(1.472\pm0.019)*10^{-1}$ & $(9.381\pm0.287)*10^{-2}$ & $1.570\pm0.047$ \\
            CFLP 200-100 & $(2.891\pm0.123)*10^{-1}$ & $(1.958\pm0.083)*10^{-1}$ & $1.477\pm0.031$ \\
            CFLP 400-100 & $(6.093\pm0.058)*10^{-1}$ & $(3.818\pm0.065)*10^{-1}$ & $1.596\pm0.024$ \\
            \hline
            CAuction 100-500  & $(1.407\pm0.095)*10^{-2}$ & $(2.983\pm0.224)*10^{-3}$ & $4.720\pm0.098$ \\
            CAuction 200-1000 & $(1.988\pm0.081)*10^{-2}$ & $(6.327\pm0.308)*10^{-3}$ & $3.144\pm0.050$ \\
            CAuction 300-1500 & $(2.484\pm0.021)*10^{-2}$ & $(9.486\pm0.152)*10^{-3}$ & $2.619\pm0.030$ \\
            \hline
            IndependentSet 500  & $(1.371\pm0.019)*10^{-1}$ & $(7.254\pm0.090)*10^{-2}$ & $1.890\pm0.018$ \\
            IndependentSet 1000 & $(5.268\pm0.047)*10^{-1}$ & $(3.117\pm0.032)*10^{-1}$ & $1.690\pm0.016$ \\
            IndependentSet 1500 & $1.305\pm0.008$           & $(7.088\pm0.046)*10^{-1}$ & $1.842\pm0.012$ \\
            \hline
            SetCover 500-1000  & $(9.406\pm4.883)*10^{-3}$ & $(2.836\pm1.678)*10^{-3}$ & $3.435\pm1.160$ \\
            SetCover 1000-1000 & $(1.851\pm0.432)*10^{-2}$ & $(7.153\pm2.052)*10^{-3}$ & $2.673\pm0.374$ \\
            SetCover 2000-1000 & $(2.791\pm0.399)*10^{-2}$ & $(1.422\pm0.267)*10^{-2}$ & $1.985\pm0.145$ \\
            \hline
        \end{tabular}
        \caption{Comparison of execution time for \inline{NodeBipartite} with cache.}
        \label{table:nodebipartite-cache}
        \end{table}

        As can be seen, Ecole is systematically faster than the literature.
        The shorter execution times in Ecole can be explained by the fact that the code from \citep{gasse2019exact} was a proof of concept and was not extensively optimized.

\section{Conclusions and Future Work}
\label{sec:future}
    Ecole offers researchers an efficient and well designed interface to the SCIP solver without compromising on customizability. 

    A current limitation of the library is that only SCIP is supported as a back-end solver.
    Since commercial, closed-source solvers such as CPLEX \citep{cplex} and Gurobi \citep{gurobi} are very popular in
    industry, it would be natural to extend the library to support them as potential back-ends.
    For now, their closed-source nature limits this possibility, but we hope that interest in the current version of Ecole
    will lead solver developers to facilitate interfacing with \gls{ml} libraries in the future.

    Future work on Ecole will involve developing new environments, such as node selection and cutting planes selection,
    as well as new observation and reward functions, such as primal, dual and primal-dual integral metrics.
    In addition, support for out-of-the-box parallelism would be useful to cope with computationally expensive environments.

\section*{Acknowledgments}
    This work was supported by the Canada Excellence Research Chair (CERC) in ``Data Science for
    Real-Time Decision Making", Mila - Quebec Artificial Intelligence Institute, and IVADO - Institut de valorisation des données.
    We are grateful to the SCIP team at the Zuse Institute Berlin for their support in developing Ecole.

\bibliography{ref.bib}

\end{document}